\documentclass[sigconf, screen]{acmart}
\AtBeginDocument{%
  \providecommand\BibTeX{{%
    \normalfont B\kern-0.5em{\scshape i\kern-0.25em b}\kern-0.8em\TeX}}}

\setcopyright{rightsretained}
\copyrightyear{2021}
\acmYear{2021}
\acmConference[DIS '21]{Designing Interactive Systems Conference
2021}{June 28-July 2, 2021}{Virtual Event, USA}
\acmBooktitle{Designing Interactive Systems Conference 2021 (DIS '21),
June 28-July 2, 2021, Virtual Event, USA}
\acmDOI{10.1145/3461778.3462115}
\acmISBN{978-1-4503-8476-6/21/06}

\newcommand{\rev}[1] {{#1}}
\begin{document}

\title{Learning Personal Style from Few Examples}

% \author{Anonymous Author(s)}
\author{David Chuan-En Lin}
\affiliation{%
  \institution{HCI Institute\\Carnegie Mellon University}
  \streetaddress{5000 Forbes Ave.}
  \city{Pittsburgh, PA}
  \country{USA}
  }
\email{chuanenl@cs.cmu.edu}

\author{Nikolas Martelaro}
\affiliation{%
  \institution{HCI Institute\\Carnegie Mellon University}
  \streetaddress{5000 Forbes Ave.}
  \city{Pittsburgh, PA}
  \country{USA}
  }
\email{nikmart@cmu.edu}

\renewcommand{\shortauthors}{David Chuan-En Lin and Nikolas Martelaro}

\begin{abstract}
  A key task in design work is grasping the client's implicit tastes. Designers often do this based on a set of examples from the client. However, recognizing a common pattern among many intertwining variables such as color, texture, and layout and synthesizing them into a composite preference can be challenging. In this paper, we leverage the pattern recognition capability of computational models to aid in this task. We offer a set of principles for computationally learning personal style. The principles are manifested in PseudoClient, a deep learning framework that learns a computational model for personal graphic design style from only a handful of examples. In several experiments, we found that PseudoClient achieves a 79.40\% accuracy with only five positive and negative examples, outperforming several alternative methods. Finally, we discuss how PseudoClient can be utilized as a building block to support the development of future design applications.
\end{abstract}

\begin{CCSXML}
<ccs2012>
   <concept>
       <concept_id>10003120.10003121</concept_id>
       <concept_desc>Human-centered computing~Human computer interaction (HCI)</concept_desc>
       <concept_significance>500</concept_significance>
       </concept>
   <concept>
       <concept_id>10010405.10010469</concept_id>
       <concept_desc>Applied computing~Arts and humanities</concept_desc>
       <concept_significance>500</concept_significance>
       </concept>
 </ccs2012>
\end{CCSXML}

\ccsdesc[500]{Human-centered computing~Human computer interaction (HCI)}
\ccsdesc[500]{Applied computing~Arts and humanities}

\keywords{style, personal preference, graphic design, machine learning}

\begin{teaserfigure}
  \includegraphics[width=\textwidth]{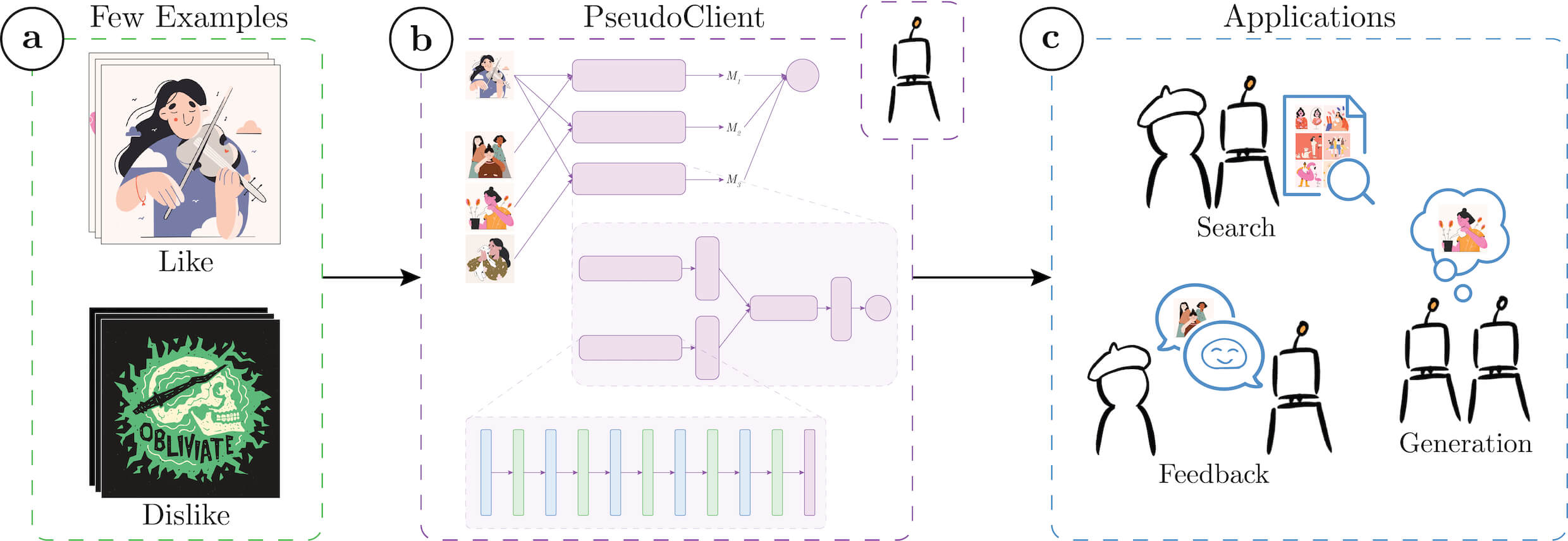}
  \caption{Given a few design examples (a), PseudoClient learns a computational model of the client's personal style preferences (b) to support multiple practical design applications (c).}
  \Description{An overview of PseudoClient's workflow}
  \label{fig:teaser}
\end{teaserfigure}

\maketitle

\section{Introduction}
A key role of the designer is being able to develop a firm grasp of the client's implicit tastes \cite{brand-personality}. For example, in the case of graphic design, a hotel group may favor a minimal and luxurious feel, while an outdoor apparel company may prefer a more bold and rugged touch. However, since clients are rarely trained to constructively articulate their design ideas, their comments may often be vague and difficult to implement \cite{voyant}.
\vspace{0.3cm}
\begin{quotation}
\textit{``Make it pop... is the dumbest thing you could say to a graphic designer.''}
\vspace{0.1cm}
\par\raggedleft--- \textup{Reddit user HylianHandy}
\end{quotation}
\vspace{0.3cm}
To gain an understanding of the client's tastes, many designers arrange early meetings with clients to probe into their personal style preferences. Some designers use specialized tools to help with this process of extracting implicit information from the clients, such as Brand Deck \cite{brand-deck}. Brand Deck consists of 100 adjective cards (e.g., vibrant, futuristic, historic) that the designer asks the client to go through and sort into three piles: you are, you are not, and does not apply. One caveat of using a tool like Brand Deck is that designers will need to further examine whether the client's understanding of subjective words matches with theirs. Other designers prefer to give clients more visual examples, such as a logo book \cite{draplin} or a mood board \cite{moodboard} consisting of a collage of different visual samples, and asking the client to pick a few samples that resonate with them. By using a selection of images, clients won't need to articulate their preferences through subjective words, but rather, by simply picking what they visually prefer. However, even when the client provides a pool of examples, recognizing a common pattern among numerous variables such as color, texture, and layout, synthesizing them into a composite understanding of the client's style preference, and using this understanding to generate a new design is a non-trivial task \cite{art-critiques}.
In this paper, we ask: can computational systems be used to aid in the task of learning peoples' style preferences from a set of example images?

Motivated by findings that high-level perceptions of a design are correlated with low-level features of appearance attributes \cite{correlating}, recent works seek to identify such low-level features using a computational approach to assist designers. Several works have explored extracting the underlying structure of designs such as the Document Object Model (DOM) tree of a web page \cite{webzeitgeist, url2video}. However, such methods only work with a small subset of design categories, such as webpages or vector images, where there is an encoded and explicit structure. For graphic design, where only bitmap information is available instead of structural elements such as shapes and text, other research has explored engineering various hand-crafted image features such as color histograms and edge maps \cite{image-style} for predicting visual style. However, akin to the limitations of rule-based AI, hand-crafted features are limited in representation power and generalizability as some appearance attributes may be too complex or abstract to manually define. More recent works have experimented with using neural networks to extract a model of visual style due to their strong ability in automatically and adaptively learning the most relevant low-level features \cite{swire, graphic-design}. In our work, we look to use neural networks to automatically learn relevant features, but with two key distinctions. First, unlike most data-hungry machine-learning-based methods, our method only needs a small handful of examples. Second, prior work mostly attempts to categorize content into subjective style terms (e.g., futuristic, minimalist) \cite{graphic-design, ui-semantics} based on the preferences of the crowd. Conversely, our work focuses on learning an individual's \textit{personal} style preference, without ever having to put a word on it nor relying on generalized conceptions of design style.
 
We introduce a set of principles for learning a client's personal style from few examples. We then use these principles to develop PseudoClient, a deep learning framework that takes in a handful of examples and automatically learns a computational model of the client's personal style. Our use case focuses on graphic design, where we feed graphic design samples as input. We first ask the client to select a few examples they like and dislike. Using the samples provided, PseudoClient learns the client's personal style preferences using a metric learning approach \cite{metric-learning} with twin Convolutional Neural Networks \cite{siamese}. Exploiting the benefits of metric learning with deep neural networks, PseudoClient requires only a few examples, does not rely on limited hand-crafted features, and can be retrained quickly with additional samples. We conduct experiments to assess how PseudoClient compares to other methods as well as how it may be affected by various factors such as training sample size and the ratio of positive and negative samples. We find that PseudoClient outperforms our baseline methods and can be tailored for different needs. Finally, we discuss PseudoClient's capability of supporting the development of future design tools in three directions: search, feedback, and generation.

In summary, our contributions are three-fold:
\begin{itemize}
\item {A set of principles for learning personal style from few examples. The principles are manifested in PseudoClient, \rev{a novel deep learning framework for learning an individual's graphic design style from a handful of examples and without relying on generalized conceptions of subjective style.}}
\item{Quantitative and qualitative experimental results demonstrating PseudoClient's advantages compared to other methods. \rev{We further examine how number of examples and ratio of positive and negative examples affect performance}.

}
\item{A discussion of potential applications where PseudoClient can support augmenting designers' capabilities. We illustrate how PseudoClient is useful as a building block for multiple practical design applications by exploring three areas of design work.}
\end{itemize}

\section{Related Work}
Our work is situated among literature in computational design understanding. More specifically, our research builds on prior work in two main branches: assessing aesthetics as a quality indicator and more fine-grained understanding of artistic style \cite{aesthetics-style}.

\subsection{Assessing Aesthetic Quality}
Computational methods of assessing the aesthetic quality of designs have a wide range of applications spanning from media editing to curation. Such methods typically involve defining a set of descriptors for some content and using them to predict human-labeled aesthetic ratings. Michailidou et al. \cite{web-aesthetic} analyzes statistics of web page elements such as number of images, words, and links and Reinecke et al. \cite{web-quantification} further incorporates page-level statistics such as number of leaves in a quadtree decomposition. In applications where the underlying structural information of the design is not available, such as bitmap photographs, prior works have engineered various visual features. Datta et al. \cite{aesthetics-photographic} extracts 56 features based on photography concepts such as rule-of-thirds and depth-of-field. Isola et al. \cite{memorable} investigates the correlation between image memorability and a set of high-level visual features such as object and scene semantics.

However, handcrafted features are nevertheless limited as some important aspects of a design may be too elusive or complex to manually define. Therefore, work in recent years has investigated the use of neural networks to automatically learn the most relevant low-level features for pattern finding. For example, NIMA \cite{nima}, one of the top performers on the Aesthetic Visual Analysis dataset \cite{ava}, explores using CNNs to automatically discover important features without human supervision. In our work, we also exploit the strength of neural networks in feature extraction and do not use manual feature engineering. Rather than predicting an aesthetics rating, we predict a match score: how well does a design align with an individual's \textit{personal} preferences as opposed to the weighted average of preferences from the crowd.

\subsection{Understanding Style}
Closely related to evaluating aesthetic quality is the task of understanding artistic style. We loosely categorize prior work in this domain into two approaches: tagging, where the task is to automatically assign labels to some content, and similarity, where the task is to learn the similarity across different content. Note that the two approaches are not necessarily mutually-exclusive.

Tagging has been extensively studied in information retrieval and recommender systems communities \cite{image-retrieval} although fewer works have explicitly approached tagging in terms of artistic \textit{style}. Prior work in tagging by style has largely treated it as some form of classification problem. Shamir et al. \cite{schools-of-art} assigns paintings to painters and schools of art using image descriptors such as color histograms and edge statistics. Karayev et al. \cite{image-style} predicts style labels such as bright and energetic for photographs and paintings using features such as color histogram and visual saliency. Wu et al. \cite{mobile-ui} models the brand personalities of mobile UIs with UI descriptors such as color, organization, and texture. \rev{Vaccaro et al. \cite{elements-of-fashion} predicts high-level fashion styles attributes such as tropical and exotic from low-level design element language such as color and material using polylingual topic modeling.} More recent works have explored style tagging with deep neural networks. Zhao et al. \cite{graphic-design} characterizes personalities for graphic designs such as cute and mysterious. \rev{Takagi et al. \cite{what-style} proposes an expert-curated fashion style dataset containing 14 categories such as rock and street and found that modern computer vision classification networks are able to outperform fashion-naïve users but not fashion-savvy users.} However, since style tags are intrinsically subjective, labeling content with style tags is by nature a noisy task. Thus, our work does not aim to describe personal style with subjective style terms. Rather, our tags are simply whether a design ``fits'' or ``does not fit'' with an individual's personal style.

The question of ``does a design fit or not fit with an individual's personal style?'' can also be reformulated as ``how similar is a design when compared to the individual's personal style?'' This formulation opens up an interesting repertoire of research in ``similarity by style'' to draw inspiration from. D.Tour \cite{dtour} and Webzeitgeist \cite{webzeitgeist} allow search by stylistic similarity based on features of a web page such as DOM tree depth and number of leaf nodes. Several other works on learning similarity by style include applications in infographics \cite{similarity-infographics}, illustrations \cite{similarity-illustration}, icons \cite{similarity-icons}, fonts \cite{similarity-fonts}, \rev{fashion \cite{128-floats}}, and 3D furniture models \cite{similarity-furniture}. In our work, we borrow from the concept of similarity by style to define personal style based on similarity with a set of representative examples selected by the individual.
\rev{Prior works have explored various methods for computing similarity, such as through color histograms \cite{color-histogram} and Convolutional Neural Networks \cite{Wang_2014_CVPR} (also see \nameref{section:comparison-with-baselines} subsection).
Recent works in the Machine Learning community have shown the effectiveness of metric learning approaches \cite{metric-learning, siamese-image-matching} for areas such as fashion \cite{Veit_2015_ICCV} and home goods product design \cite{bell2015learning}.
We build upon the successes of metric learning to design a metric learning framework for personal graphic design style.
To guide our work, we distill a set of principles.}

\section{Principles of Learning Personal Style}
To support our objective of learning personal style preferences, we ground the development of PseudoClient in four central principles.

\subsection{Principle 1: Learn by Example}
Considering that design vocabulary (e.g., minimal, vintage) is inherently subjective and highly dependent and interpreted based on the individual's knowledge and experiences \cite{aesthetics-style}, we do not ask clients to indicate their preferences through such vocabulary nor do we attempt to fit their preferences into such vocabulary. 
% Afterall, there are $n = number\:of\:people\:in\:the\:world$ categories of aesthetic style \cite{aesthetics-wiki}. 
Inspired by the mood board technique \cite{moodboard}, we simply ask the client to supply us with visual examples and learn to judge the client's personal style based on them.

\subsection{Principle 2: Learn by a Handful}
Since our task is to learn personal style, our examples come directly from the client of interest. However, asking the client to select massive amounts of examples is tedious. In addition, prior works have found that as the quantity of examples required from the client increases, the quality and consistency of the examples begin to decrease due to fatigue and the pressure to reach the target of providing a high number of samples \cite{label-accuracy}. Based on these observations, rather than being data-hungry, PseudoClient should be capable of working with only a handful of examples.

\subsection{Principle 3: Learn by Juxtaposition}
Findings from prior work in recommender systems \cite{recommender-systems} suggest that it is easier to select positive and negative samples than to discern likeability on a spectrum. Given this, we ask the client to provide us a set of examples they like (positive samples) and another set of examples they dislike (negative samples). Our task then is to determine whether a design fits the positive samples or the negative samples more closely. With only a limited number of examples to learn from, learning by juxtaposition allows us to formulate our problem of modeling the client's style preferences into more simplistic binary classification problems.

\subsection{Principle 4: Learn by Multiple Comparisons}
Studies in the learning sciences and cognitive psychology have observed that through comparison against multiple examples, one can quickly learn a common underlying structure, even if the individual examples are not fully understood \cite{schema,analogical-encoding}. Building on this insight, we repeatedly do pairwise comparisons between the unseen design and the pool of reference examples provided by the client. For a more detailed description and diagram of the method, please refer to the \nameref{section:comparison-framework} subsection.

\begin{figure*}[htbp]
  \centering
  \includegraphics[width=12.5cm]{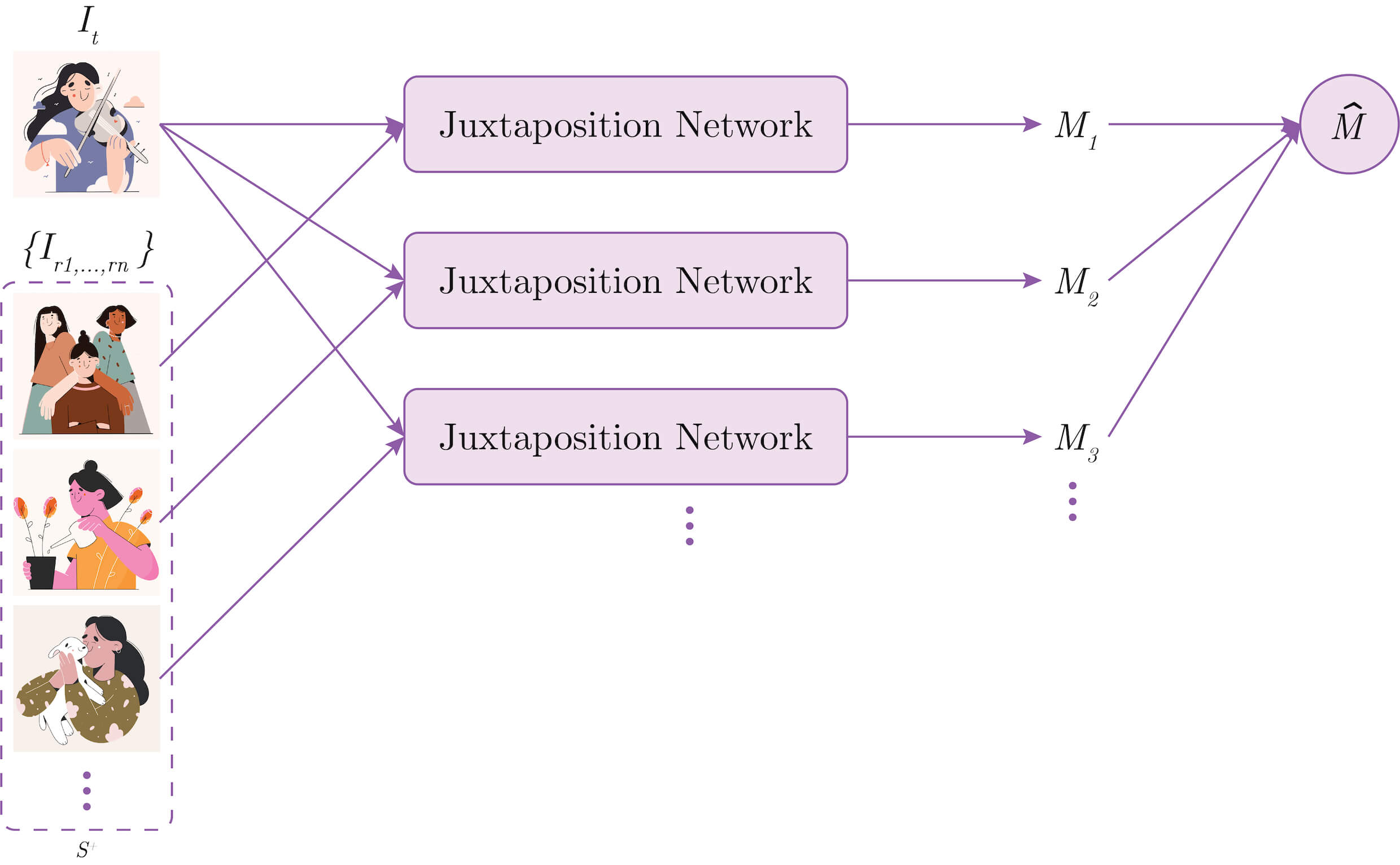}
  \caption{The Comparison Framework. We compute pairwise match scores between the test image and each reference image in the positive support set. The final match score is the median.}
  \Description{Diagram of the Comparison Framework}
  \label{fig:comparison-framework}
\end{figure*}

\section{Implementation}
Our four principles are manifested in PseudoClient and guide its implementation. \rev{To the best of our knowledge, we uniquely frame our task of modeling the client's \textit{personal graphic design style} as a metric learning problem \cite{metric-learning}.
Given an unseen graphic design, our objective is to determine its \textit{similarity} with a set of representative examples selected by the client.} This allows us to then classify the design between two classes: client likes and client dislikes. The following outlines the implementation of PseudoClient, including (1) the Support Set, (2) the Comparison Framework, (3) the Juxtaposition Network, (4) the Embedding Network, and (5) the training setup. We wrote our learning framework in PyTorch version 1.6.0.

\subsection{Support Set}
We take inspiration from the mood board technique \cite{moodboard} by first asking the client to provide a small selection (Principle 2) of graphic design examples (Principle 1). More specifically, we ask the client to pick a few samples they like (positive samples) and another few samples they dislike (negative samples) (Principle 3). This is our support set $S$. We denote a subset of $S$ containing only positive samples as the positive support set $S^{+}$ and a subset of $S$ containing only negative samples as the negative support set $S^{-}$.

\subsection{Comparison Framework}
\label{section:comparison-framework}

Figure~\ref{fig:comparison-framework} visualizes the Comparison Framework. Given an unseen graphic design $I_{t}$ and a positive support set $S^{+}$, we perform pairwise comparisons (Principle 4) between $I_{t}$ and each reference sample $I_{r1,...,rn} \in S^{+}$ to compute their respective match scores $M_{1,...,n}$ through our Juxtaposition Network (see \nameref{section:juxtaposition-network} subsection on how $M$ is computed).
A high match score means that $I_{t}$ is predicted to be in the same class as $I_{r}$. This implies that the unseen design $I_{t}$ matches the client's personal style, since $I_{r}$ is a sample that the client likes. Conversely, a low match score means that $I_{t}$ is predicted to be in a different class as $I_{r}$. This implies that the unseen design $I_{t}$ does not match the client's personal style. We are therefore learning to classify designs \textit{indirectly} by evaluating their similarities with a set of labeled referencing examples (positive support set $S^{+}$).
\rev{We compute match scores with the positive support set $S^{+}$ as opposed to the full support set $S$ based on our findings from empirical pilot testing.
We found that the client's negative examples tend to be less consistent, often consisting of many diverging styles the client dislikes.}
We then take the median of the pairwise match scores $M_{1,...,n}$ as the overall predicted match score $\hat{M}$. We take the median rather than the mean to decrease the effects of outliers.
\begin{displaymath}
  \hat{M}(I_{t}) = \mbox{Median}_{I_{r} \in S^{+}}(M(I_{t}, I_{r}))
\end{displaymath}

\subsection{Juxtaposition Network}
\label{section:juxtaposition-network}
Since we want to work with a small support set (Principle 2), we approach the challenge of learning a model to accurately predict the match score $M$ as a few-shot learning task \cite{one-shot}. \rev{Prior work by Melekhov et al. \cite{siamese-image-matching} in image matching showed the strength of Siamese Networks \cite{siamese} in generalizing from small datasets. We build upon this and learn the similarity function between two graphic design inputs. We design a Juxtaposition Network resembling a Siamese Network with twin Convolutional Neural Networks (CNN).}

\begin{figure*}[htbp]
  \centering
  \includegraphics[width=15cm]{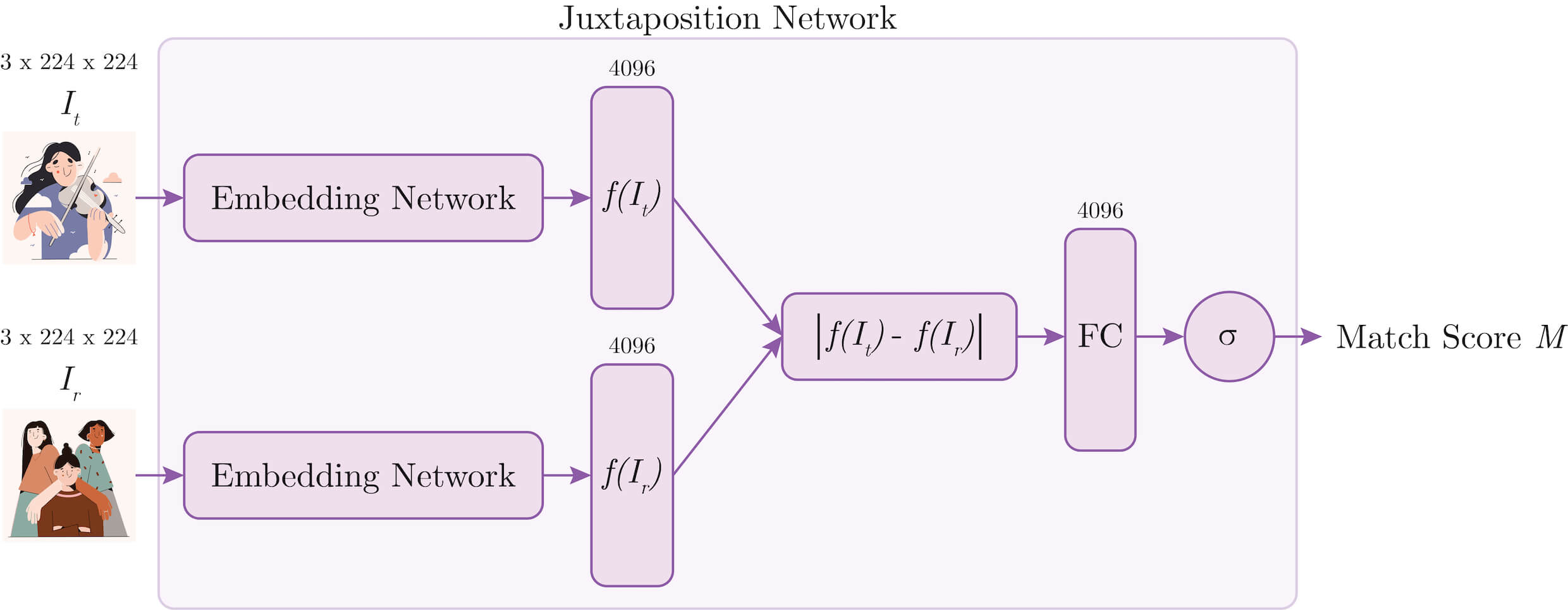}
  \caption{Architecture of the Juxtaposition Network}
  \Description{Architecture of the Juxtaposition Network}
  \label{fig:juxtaposition-network}
\end{figure*}

\begin{figure*}[htbp]
  \centering
  \includegraphics[width=\linewidth]{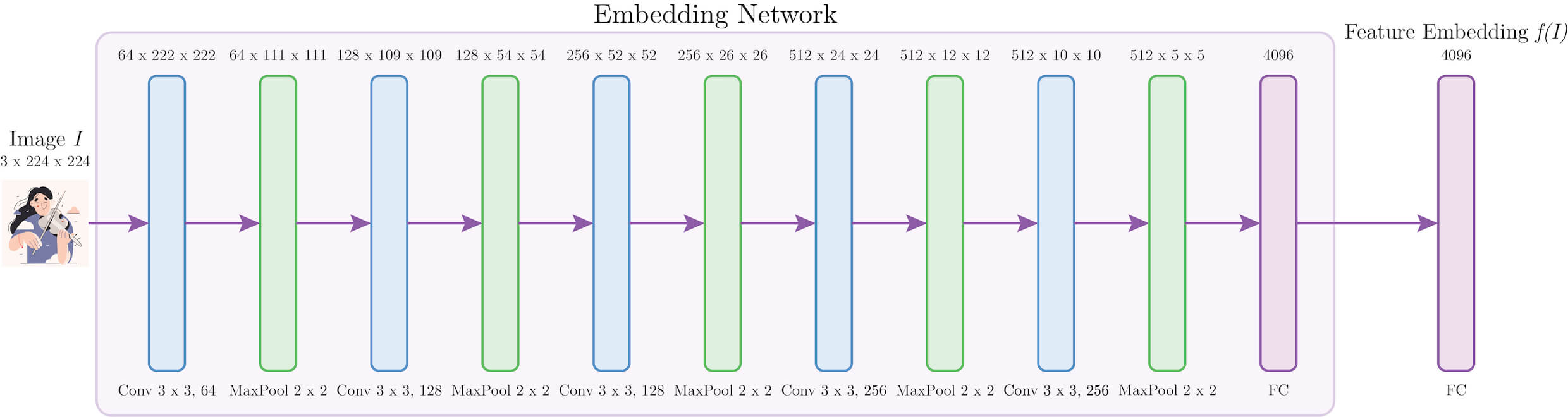}
  \caption{Architecture of the Embedding Network}
  \Description{Architecture of the Embedding Network}
  \label{fig:embedding-network}
\end{figure*}

Figure~\ref{fig:juxtaposition-network} visualizes the architecture of the Juxtaposition Network. The Juxtaposition Network takes in a test image $I_{t}$ and a reference image $I_{r} \in S^{+}$ as inputs and computes a match score $M$ with $range=[0, 1]$ as output. Our Juxtaposition Network uses a twin architecture. We first encode each 3x224x224 RGB input image $I_{t}$ and $I_{r}$ into 4096-dimensional feature embeddings $f(I_{t})$ and $f(I_{r})$ through twin Embedding Networks (see \nameref{section:embedding-network} subsection). We then compute the weighted L1-distance $D$ between the two feature embeddings.
\begin{displaymath}
  D(I_{t}, I_{r}) = |f(I_{t}) - f(I_{r})|
\end{displaymath}
We then translate the distance $D$ into a match score $M$, which is the probability that the two inputs belong to the same class, by passing it through a fully-connected layer (FC) with learned weights $\mathcal{W}$ and a sigmoid activation function ($\sigma$).
\begin{displaymath}
  M(I_{t}, I_{r}) = \frac{1}{1 + \exp^{-\mathcal{W}\cdot D(I_{t}, I_{r})}}
\end{displaymath}

\subsection{Embedding Network}
\label{section:embedding-network}

Figure~\ref{fig:embedding-network} visualizes the architecture of the Embedding Network. The Embedding Network takes in a 3x224x224 RGB image as input and extracts a 4096-dimensional feature embedding as output. The feature embedding is a vector representation that captures visual features of the image. We use five convolutional blocks (a 3x3 convolutional layer and a 2x2 max-pooling layer) and two fully-connected layers. We apply batch normalization and ReLU activation after the convolutional layers and sigmoid activation after the fully-connected layers. Our architectural design resembles the well-researched VGG architecture \cite{vgg} and performs well empirically. While there are certainly techniques to further optimize performance \cite{architecture-search}, they are not the focus of this paper.

\begin{figure*}[htbp]
  \centering
  \includegraphics[width=15cm]{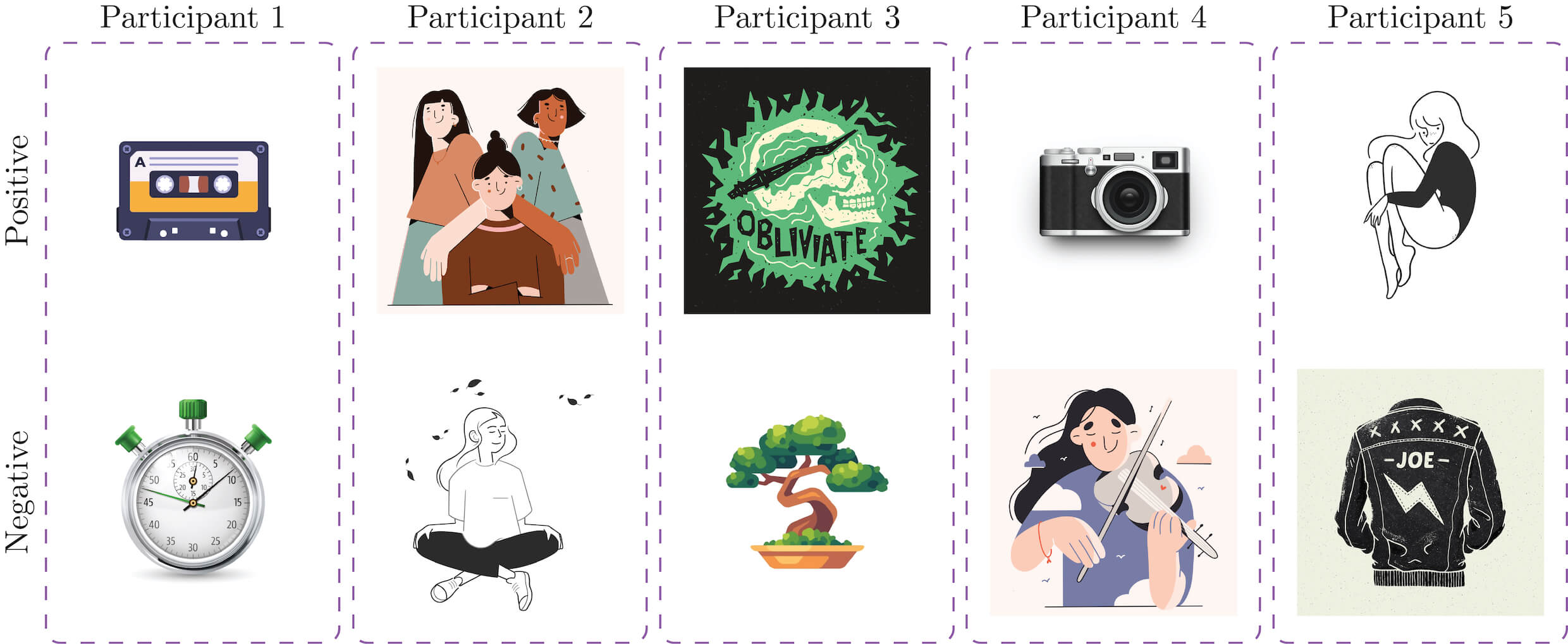}
  \caption{A snapshot of each participants' positive and negative sets. Each column represents a participant. The top row displays an example of their positive set and the bottom row displays an example of their negative set.}
  \Description{Some sample data for the experiments}
  \label{fig:snapshot}
\end{figure*}

\subsection{Training Setup}
Our training setup consists of two stages: pre-training and finetuning.

\subsubsection{Pre-training}
To allow our model to quickly learn a client's personal style with few examples, we first pre-train a network on a dataset of graphic design images collected from \href{https://dribbble.com/}{dribbble.com}. The dataset consists of six different classes (flat, geometric, line, minimal, pop, vintage) with 50 images each, yielding a collection of 300 images. We collected the dataset based on querying relevant keywords, with class labels self-tagged by the artists. Note that the main purpose of pre-training is to \rev{seed the network}, letting our model first learn some basic visual features of graphic design, such as edges, patterns, or general shapes, \rev{an interesting property of neural networks \cite{visualizing}}. Thus, our focus is \textit{not} to bin all graphic design work into these six subjective categories. Rather, when the client trains PseudoClient to model their own tastes, instead of training a model from scratch with completely random initialized weights, the client's model can have a head start by building on the pre-trained model via transfer learning \cite{transfer-learning}, resulting in a much shorter training time.

We process the dataset into \textit{pairs} of graphic design images with an assigned binary label $y$. If the images in a pair belong to the same class, we set $y = 1$. If the images in a pair belong to distinct classes, we set $y = 0$. The pairs are randomly sampled. We resize the images into 3x224x224 pixels with RGB channels and normalize each color channel with $mean=[0.485, 0.456, 0.406]$ and $std=[0.229, 0.224, 0.225]$ based on the statistical distribution of ImageNet \cite{imagenet}. We split our training and validation sets with a 9:1 ratio. We use the Adam optimizer \cite{adam}, a batch size of 32, a learning rate of $1 \times 10^{-5}$, and train for 50 epochs. Since the labels are binary, we use a cross-entropy loss $\mathcal{L}$ for every training batch $B$.
\begin{displaymath}
  \mathcal{L}(B) = \sum_{I_{t}, I_{r}, y \in B}y\log(M(I_{t}, I_{r})) + (1-y)\log(1-M(I_{t}, I_{r}))
\end{displaymath}
Training takes around an hour to complete on an NVIDIA GeForce RTX 2080 Ti graphics card with 11GB of memory.

\subsubsection{Fine-tuning}
This is the stage where the model learns the client's \textit{personal} preferences. Given a pre-trained model, we fine-tune the model so that it reflects their own tastes. We begin with the pre-trained weights and train the model with respect to their support set $S$ as the new training dataset. Data processing procedures are the same as for the pre-trained model. Therefore, we learn by juxtaposition (Principle 3), where $y = 1$ means that the graphic designs in the pair are both liked or disliked by the client (same class) and $y = 0$ means that one of the graphic designs in the pair is liked by the client while the other is disliked by the client (different class). This means we are not directly learning whether a design is liked or disliked by the client, but rather learning its similarity with both the liked and disliked example sets. We train our model until it can consistently predict the correct $y$ label for each pair. We use the same optimizer and loss function as the pre-trained model, a batch size of 16, a learning rate of $1 \times 10^{-8}$, and train for 20 epochs. Training takes around ten minutes to complete on the same hardware as for the pre-trained model.

\section{Experiments}
To evaluate the effectiveness of PseudoClient, we performed several experiments. \rev{This serves as a litmus test of the system's performance.} The following outlines our setup and various experiments, including:
\begin{itemize}
\item {\textbf{Comparisons with baselines.} We evaluate PseudoClient's performance by comparing against other methods, namely, softmax-based approaches (Convolutional Neural Networks) and traditional distance measurements (color histogram distance).}
\item{\textbf{Exploration of various factors.} To discover usage guidelines for future designers and offer an understanding of how designers can work with PseudoClient for their needs, we explore how various factors can affect PseudoClient's ability to learn personal style. We focused on two factors: number of examples (dataset size) and different ratios of positive and negative examples (class imbalance).}
\item{\textbf{Query results.} Finally, as graphic design is a highly visual medium, we want to see how PseudoClient performs qualitatively to uncover insights that may not be portrayed through numbers. We do this through a simple image retrieval task, visualizing the retrieval results.}
\end{itemize}

\begin{table*}[t]
  \caption{The accuracies of different methods between participants P1 -- P5 and overall. The highest accuracy is highlighted in bold.}
  \label{table:comparison}
  \resizebox{450pt}{!}{
  \begin{tabular}{ccccccc}
    \toprule
    Method&P1 Accuracy&P2 Accuracy&P3 Accuracy&P4 Accuracy&P5 Accuracy&Overall Accuracy\\
    \midrule
    \textit{Chance}&\textit{50.00\%}&\textit{50.00\%}&\textit{50.00\%}&\textit{50.00\%}&\textit{50.00\%}&\textit{50.00\%}\\
    Color Histogram&69.00\%&26.00\%&53.00\%&55.00\%&58.00\%&52.20\%\\
    CNN&67.00\%&63.00\%&76.00\%&53.00\%&62.00\%&64.20\%\\
    PseudoClient (Ours)&\textbf{86.00\%}&\textbf{77.00\%}&\textbf{90.00\%}&\textbf{64.00\%}&\textbf{80.00\%}&\textbf{79.40\%}\\
    \bottomrule
  \end{tabular}
  }
\end{table*}

\subsection{Setup}
\label{section:setup}
Given that our task is to learn personal style preferences, we evaluate our accuracy in doing so with five people: two of the paper's authors and three volunteers. For each participant, we ask them to first come up with a simple design idea (e.g., an illustration for a local French bakery). We then ask them to select 70 images that fit their design idea (positive set) from \href{https://dribbble.com/}{dribbble.com}. This is for our experimental purposes, such that we have enough images for various levels of training and testing. In an actual usage scenario, participants would not need to supply as many examples (see \nameref{section:number-of-examples} subsubsection).

After all participants finish selecting their positive sets, participants then select another person's positive set (from the pool of positive sets of all other participants) as their negative set. Our requirement is that their selected negative set doesn't fit their design idea. One potential constraint of this approach is that a participant may struggle in selecting a negative set from the pool if all other positive sets match closely with theirs. However, such an issue did not occur during our specific study (maybe because participants had different design ideas). Thus, each positive set was stylistically different from the negative set. For a snapshot of what the participants picked as their positive and negative sets, please refer to Figure~\ref{fig:snapshot}. For each positive and negative set of 70 images, we randomly allocate 50 as our \textit{test} set and 20 as our \textit{training} set. This means that 50 of the 70 images will be used as a true test set and not be used for training. For each training set of 20 examples, we further randomly sample various training sets of 1, 5, 10, and 20 examples for our later study on how different numbers of training samples affect performance (see \nameref{section:exploration} subsection). For our comparison with baselines and qualitative query results studies, we use the training set of 5 examples (i.e., training with only 5 positive and negative examples). This follows our objective of learning by a \textit{handful} of examples (Principle 2).

We train a separate model for each participant. To measure the accuracy of the model, we then compute a match score for each of the 50 examples in their unseen positive test set as well as for each of the 50 examples in their unseen negative test set, by pairwise comparisons with all examples in their positive training set (see \nameref{section:comparison-framework} subsection). Note that the match score $M$ is a numeric value with $range = [0, 1]$ where a higher value implies higher predicted similarity with the positive support set, and a lower value implies a lower predicted similarity with the positive support set. For the positive test set, if $M$ is greater than the threshold of 0.5, we note down a correct true positive ($TP$) prediction. For the negative test set, if $M$ is less than the threshold of 0.5, we note down a correct true negative ($TN$) prediction. Our accuracy is then given by $\frac{number\:of\:TP\:+\:number\:of\:TN}{100}$. Finally, we take the average of the accuracies for each participant as the overall accuracy.

\subsection{Comparison with Baselines}
\label{section:comparison-with-baselines}
We implemented two baseline models to compare against our method: (1) color histogram and (2) a convolutional neural network (CNN). Our first baseline is based on the distance of color histograms, which is a widely used metric for evaluating similarity between images \cite{color-histogram}. We first extract a 3D histogram from each RGB image, using 8 bins per channel and normalize with $range=[0, 256]$. We then flatten the histogram, yielding a 512-dimensional vector. To compute the distance between a pair of vectors, we compute its correlation \cite{opencv}. The correlation is a numeric value with $range=[0, 1]$, where higher correlation implies smaller distance and more similarity and and lower correlation implies larger distance and less similarity. We compute the overall accuracy of the color histogram method using a similar procedure as the second paragraph of the \nameref{section:setup} subsection. However, instead of using a fixed threshold of 0.5, we set the threshold as the mean correlation value between images in the training set since the distribution of correlation values differ greatly across different participants. 

Our second baseline is a standard implementation of a CNN, representing a typical neural-network-based approach for classification. We use the architecture of our \nameref{section:embedding-network} with a single output node. Rather than computing a match score, we directly classify whether the unseen test image is liked or disliked by the participant. The accuracy is then given by $\frac{number\:of\:correct\:predictions}{100}$ and we also take the average of the accuracies for each participant as the overall accuracy.

\begin{table*}
  \caption{The accuracies of PseudoClient when given different numbers of examples between participants P1 -- P5 and overall. The highest accuracy is highlighted in bold.}
  \label{table:number}
  \resizebox{420pt}{!}{
  \begin{tabular}{ccccccc}
    \toprule
    \# Examples&P1 Accuracy&P2 Accuracy&P3 Accuracy&P4 Accuracy&P5 Accuracy&Overall Accuracy\\
    \midrule
    1&47.00\%&74.00\%&47.00\%&53.00\%&70.00\%&58.20\%\\
    5&86.00\%&77.00\%&90.00\%&64.00\%&80.00\%&79.40\%\\
    10&83.00\%&73.00\%&93.00\%&66.00\%&75.00\%&78.00\%\\
    20&\textbf{88.00\%}&\textbf{80.00\%}&\textbf{96.00\%}&\textbf{77.00\%}&\textbf{92.00\%}&\textbf{86.60\%}\\
    \bottomrule
  \end{tabular}
  }
\end{table*}

\begin{table*}
  \caption{The overall true positive percentages, true negative percentages, accuracies, and F1 scores of PseudoClient when given different ratios of positive and negative examples. The highest results are highlighted in bold.}
  \label{table:ratio}
  \resizebox{380pt}{!}{
  \begin{tabular}{ccccc}
    \toprule
    Number of Positive:Negative Examples&True Positives&True Negatives&Accuracy&F1 Score\\
    \midrule
    10:5&\textbf{82.40\%}&68.40\%&\textbf{75.40\%}&\textbf{\rev{0.77}}\\
    5:10&65.20\%&\textbf{82.80\%}&74.00\%&\rev{0.71}\\
    \bottomrule
  \end{tabular}
  }
\end{table*}

By learning to classify designs \textit{indirectly} via learning their similarities with a positive support set, our hypothesis is that PseudoClient would perform better than a standard CNN implementation, given the nature of our task: ``whether a design is more similar to the set of like examples or the set of dislike examples'' seems more learnable than naively judging ``whether a design is liked or disliked.'' The latter is an intrinsically ambiguous task, since like/dislike more often falls on a scale as opposed to being a perfect dichotomy. In addition, we also hypothesize that color alone would not be sufficient for determining personal style. From Table ~\ref{table:comparison}, we observe that color histogram performs only marginally better than random chance. The CNN, which is essentially PseudoClient's Embedding Network, still struggles to classify consistently. Overall, we observe that PseudoClient is able to outperform our two baselines for all participants, supporting our hypotheses.

\subsection{Exploration of Various Factors}
\label{section:exploration}
\subsubsection{Number of Examples}
\label{section:number-of-examples}
We first investigate how supplying different training sample sizes affects PseudoClient's performance. Our hypothesis is that accuracy will increase as the number of examples increases, and we test our hypothesis by training separate models using 5, 10, and 20 positive and negative examples and evaluating their accuracies. We do not explore beyond 20 examples since it goes beyond the definition of ``few examples'' based on feedback from our participants. We also evaluate how well our pre-trained model can generalize to personal style \textit{without} any further fine-tuning and given only 1 reference example.

Table ~\ref{table:number} summarizes the accuracies of PseudoClient when given different numbers of examples for each participant and overall. We observe that accuracy generally increases as the number of examples increases, confirming our hypothesis. However, interestingly, when compared to the overall accuracy of the models using 5 examples, the overall accuracy of the models using 10 examples did not increase and even dipped slightly. This may suggest that an increase in performance may only be prompted by a sufficiently large increase in sample size. Another interesting observation is that our pre-trained model, given only 1 guiding example, is able to perform better than random chance by some amount for P2 and P5. The reason may be that the examples chosen by these participants are more uniform, granting the single example a greater representative capacity. Hence, designers should note that clients who have tighter style preferences may not need to supply as many examples.

\subsubsection{Ratio of Positive and Negative Examples}
Previously, we trained our models with equal amounts of positive and negative examples. We thought it would be interesting to also investigate how performance would be affected if the client gives more positive examples and fewer negative examples, and vice versa. We hypothesize a higher $TP$ when given a higher ratio of positive examples and a higher $TN$ when given a higher ratio of negative examples. We experimented with two setups: 10 positive examples with 5 negative examples and 5 positive examples with 10 negative examples.

\begin{figure*}[htbp]
  \centering
  \includegraphics[width=15cm]{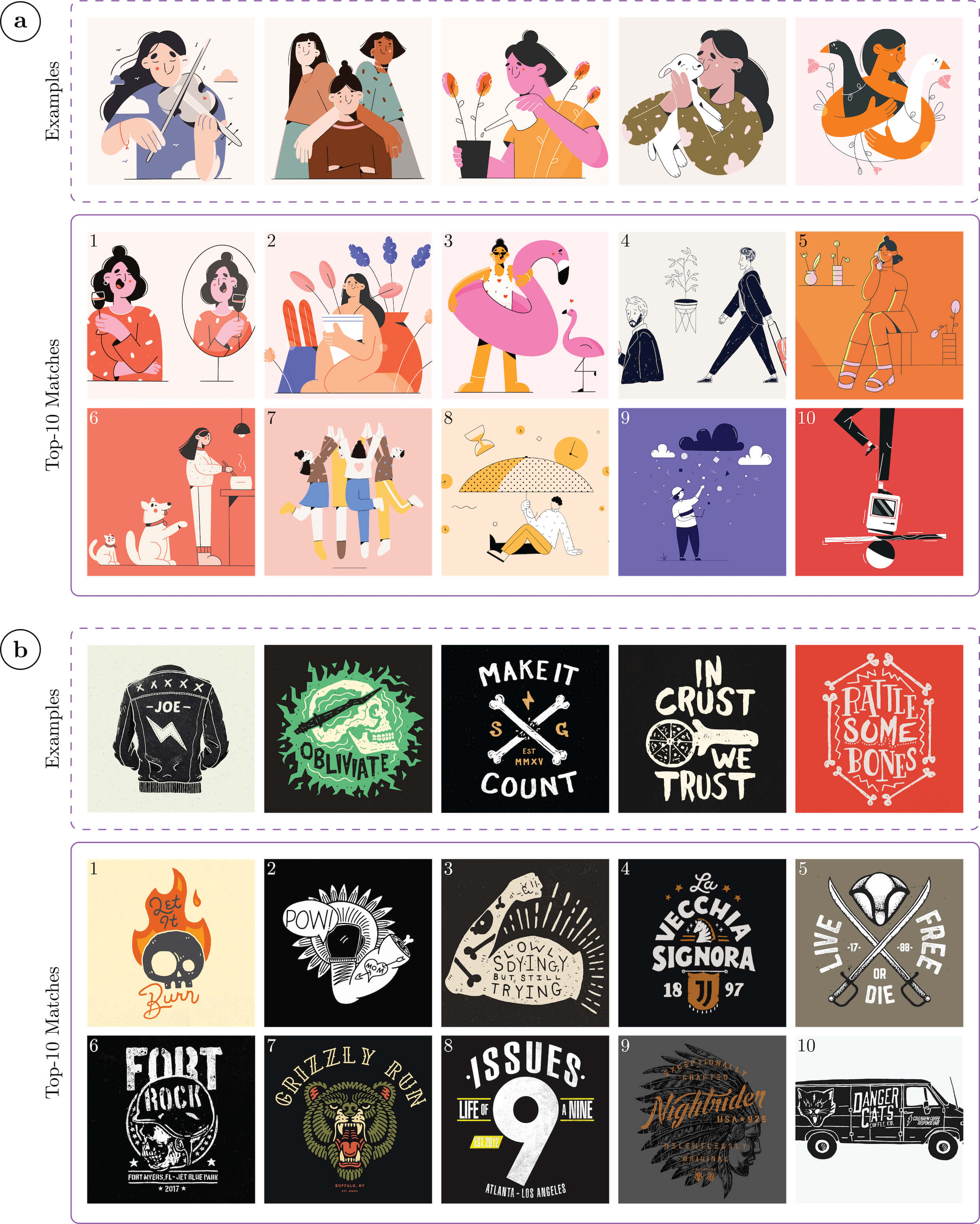}
  \caption{(a) and (b) show two example query results. The dashed lines contain the positive examples used for training (positive support set). The solid lines contain the queried samples with the highest match scores, ranked from 1 to 10.}
  \Description{Some sample query results}
  \label{fig:query}
\end{figure*}
% \clearpage

Table ~\ref{table:ratio} summarizes the overall true positive percentages, true negative percentages, accuracies\rev{, and F1 scores} of PseudoClient when given different ratios of positive and negative examples. We observe that the model predicts true positives more consistently when given more positive examples and predicts true negatives more consistently when given more negative examples, supporting our hypothesis. This reveals an interesting property: designers may adjust the ratio of positive and negative examples required for their specific goals. For example, if the goal is targeted towards identifying and filtering out what the client \textit{dislikes}, then the designer may ask the client to focus on supplying more negative examples.

\subsection{Query Results}
We qualitatively evaluate the performance of PseudoClient with an image retrieval task. We query for the top 10 images with the highest match scores from a database of graphic design samples. Our database consists of a subset of the Dribbble dataset from \cite{ui-semantics} and some examples selected by the participants. None of the samples were used for training.

Figure~\ref{fig:query} visualizes two example queries. The samples bounded by the dashed lines are the positive examples used for training (positive support set) and the samples bounded by the solid lines are the top-10 queried results ranked from 1 to 10. We observe that PseudoClient is able to retrieve stylistically similar graphic designs by synthesizing from a few examples. Interestingly, a couple retrieved designs in Figure~\ref{fig:query}a even belong to the same artist as the provided examples, demonstrating PseudoClient's capability of recognizing personal design style. Note that the retrieved samples don't necessarily have similar color as the provided examples. For example, the top-ranking retrieval in Figure~\ref{fig:query}b has a light cream background despite most of the examples having darker backgrounds. Nonetheless, its resemblance to the examples in terms of artistic style is apparent. One limitation we discovered from the queried results is that PseudoClient judges samples in a more holistic sense and may overlook fine-grained but important details such as font styling (see Figure~\ref{fig:query}b). For example, serif and san-serif fonts may elicit different emotions \cite{font-perception}. We suggest an approach to address this in the \nameref{section:limitations-future-work} section.

\begin{figure*}[htbp]
  \centering
  \includegraphics[width=15cm]{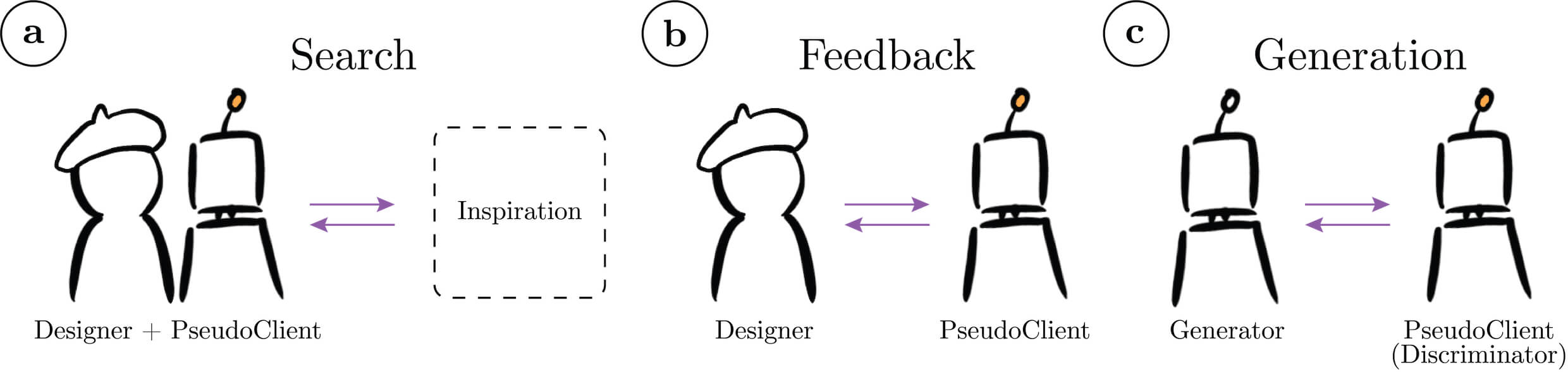}
  \caption{PseudoClient can be used as a building block for multiple practical design applications. We explore applications from three directions: (a) search, (b) feedback, and (c) generation.}
  \Description{Design space of PseudoClient}
  \label{fig:applications}
\end{figure*}

\section{Applications}
\label{section:applications}
We suggest possible applications that can be enabled using PseudoClient by illustrating its use cases in augmenting designers from three directions: (1) search, (2) feedback, and (3) generation.

\subsection{Search}
A natural application of PseudoClient is an example-based, personalized style search engine (Figure~\ref{fig:applications}a). Prior work has shown that being able to find high quality examples is a crucial part of the creative design process for not only gaining inspiration, but also exploring alternatives and performing comparative evaluations \cite{getting-inspired}. An example usage scenario may be that shown Figure~\ref{fig:query}. The designer may first request a few examples from the client. Using these examples, the designer may then search for even more examples of similar style. Note that unlike existing search tools built into many design sharing websites, a search engine built on top of PseudoClient has the benefit of searching directly with examples instead of tagging based search with subjective keywords \cite{ui-semantics}. Furthermore, style-based search can surface designs that have not been tagged with keywords. This setup provides a powerful mechanism for locating potential relevant design examples to draw inspiration from, without having to overwhelm the client with the task of providing large quantities of examples themselves or going through lengthy meetings for designers to probe and synthesize their tastes.

Extending beyond searching for design examples, PseudoClient may also be used for clients to search for \textit{designers}. Since designers often also have their own personal style, it is sensible for clients to find designers who have personal styles that match their tastes. Given a designer's design portfolio, PseudoClient can assess the portfolio's overall similarity with the client's personal style preferences. The client may thus do a ``reverse designer search'' with PseudoClient to discover the top designers who most fit their needs. Similarly, one may also do a ``reverse design community search'' with PseudoClient to discover the top design communities that are most well aligned with their personal style preferences. A fellow designer pointed out that being able to cluster designers or design communities by style may also reveal how designers are influenced by one another and unravel interesting design trends and patterns.

\subsection{Feedback}
The ability of PseudoClient to assess similarity by style opens the possibility of an automatic design feedback system. By referencing the few examples given by the client, PseudoClient can provide rapid, automatic design feedback to the designer for evaluating how well the designer's design drafts align with the client's personal tastes (Figure~\ref{fig:applications}b). Such an application can potentially increase the quality of design work by allowing the designer to receive timely critique for iterative exploration of alternatives and ensure that their design direction remains aligned with the client's tastes, without the constraint of the client's limited availability \cite{time-constraints}. This is especially timely as an increasing amount of design work shifts to nowhere and everywhere (remote) where latency between design cycles is substantially magnified due to time zone differences and the absence of face-to-face meetings.

Another interesting use case of PseudoClient's capability of giving feedback, suggested by another fellow designer, is a tutorial system for beginners to mimic the styles of masters.
\vspace{0.3cm}
\begin{quotation}
\textit{``Art is sourced. Apprentices graze in the field of culture.''}
\vspace{0.1cm}
\par\raggedleft--- \textup{Jonathan Lethem, Writer}
\end{quotation}
\vspace{0.3cm}
A common way for beginner artists to improve their techniques is by performing master studies. During this process, the beginner studies the techniques of a master by recreating a piece of work using a similar style \cite{master-copy}. However, whether the master copy truly aligns with the master's style may ultimately be difficult to determine. Given a handful of the master's work as examples, PseudoClient can be applied as a simple tutorial system by rating the beginner's master copy. Implementing an activation map to reveal which regions of the copy matches with the master's style may further increase explainability and actionability of PseudoClient's feedback \cite{visual-importance}. Such an application can also extend to mimicking the artist style of an era or genre.

Since PseudoClient effectively learns a model of one's personal style preferences, its feedback capability can also be repurposed for personal authentication. Google's reCAPTCHA system serves hundreds of millions of CAPTCHAs every day to tell humans apart from computers \cite{captcha}. Users are presented a set of 9 or 16 square images and asked to identify which images contain certain objects. For our use case, we can first ask a user to select a small positive support set of design images upon the creation of an account. PseudoClient may then serve as an authentication system, in similar fashion to reCAPTCHA, by asking the user to select the design images that most align with their personal style preferences and judging its overall stylistic similarity with the positive support set associated with the account for authentication. The core assumption behind is that one's unique style preference can be personal enough to serve as a \textit{mental metric} (as opposed to biometrics) for personal identification.

\subsection{Generation}
PseudoClient may also serve as a key component in generative design methods, such as Generative Adversarial Networks (GANs) \cite{gan}. We can think of GANs as two actors, a generator and a discriminator, working against each other to generate realistic designs. The generator attempts to hallucinate ``fake'' yet plausible designs. The discriminator attempts to differentiate between ``fake'' designs, generated by the generator, and ``real'' designs, from a set of real design examples. Competing against each other, each actor becomes better and better at doing its job, until the generated ``fake'' design is barely distinguishable from a ``real'' design. PseudoClient may be used as a discriminator (Figure~\ref{fig:applications}c). Instead of differentiating between ``real'' and ``fake'' designs, our new discriminator differentiates between designs that the client ``likes'' or ``dislikes'' with respect to their personal style preference. On the other hand, the generator (or ``PseudoDesigner'') attempts to minimize the difference between its generated designs and the ``like'' examples provided by the client. Designers may thus utilize this adversarial behavior to synthesize designs that resemble the client's personal style preferences. We hope that such a generative system can become a useful tool, not to replace the role of designers, but to assist designers in serving as a source of inspiration, for rapidly prototyping new alternatives, and fundamentally making design work more accessible to novices.

\section{Limitations and Future Work}
\label{section:limitations-future-work}
While PseudoClient performs well and appears to learn personal style, there are several limitations. 
These limitations suggest interesting avenues for future work. 
First, while PseudoClient is able to learn personal style holistically, more fine-grained elements such as variations in typeface may be overlooked\rev{, perhaps due to downsampling}. 
A possible approach to address this may be to first semantically segment a design in order to distinguish between different design elements, as opposed to treating the entire design as a whole. 
For example, a design's typeface or background (with foreground subtracted) could be treated as separate features for learning. This being said, adding more feature engineering could lead to overly constrained systems.
Second, a limitation raised by a fellow designer is that while PseudoClient is able to learn style in a \textit{visual} sense, it is not able to explicitly understand a design based its \textit{content}. For example, the use of skulls and bones in Figure~\ref{fig:query}b may correlate with specific styles (e.g., grunge, vintage, spooky), although their usage per individual may be different based on how they interpret these symbols and the context that surrounds them. 
It would be interesting to explore how content could correlate to personal elicitations of style and how (in)consistent they might be across different individuals.
Third, while our evidence suggests that PseudoClient is able to distinguish between different styles, we may see that the positive and negative styles of participants in our experiments were quite different from each other. This motivates future work on exploring various degrees of similarity between positive and negative styles, such as investigating how PseudoClient can learn very subtle style differences.
Finally, PseudoClient currently functions more as a component rather than as a fully fleshed out design application. For future work, we plan to work with PseudoClient as a design material \cite{ai-as-ux} to implement some of the design tools discussed in the \nameref{section:applications} section and evaluate them via user studies with designers and clients.

\rev{\section{Potential Risks and Implications}
The introduction of PseudoClient into the design workflow may also come with potential risks and implications. For example, due to the nature of black box models, designers may not be able to interpret and explain the suggestions made by PseudoClient to their clients.
Arguably, clients may feel that the design decisions are purely arbitrary.
An extension to increase interpretability may be to generate an activation map \cite{gradcam} that visualizes which regions of the design the model focuses on for making its predictions.
Another potential risk is that our system when utilized in a search system could bury certain artists' work due to potential biases in the network.
Future work should explore whether different decisions in the modeling process may lead to some kinds of work not being surfaced.
Ultimately, as with any assistive system in a real world deployment, one should consider how the system affects various stakeholders and be wary of being overly reliant on the system's suggestions.}

\section{Conclusion}
This paper demonstrates how we can leverage the pattern recognition capability of computational models to learn personal style preferences from only a few examples. We offer a set of principles built on prior work to ground our solution. Based on these principles, we designed PseudoClient, a metric-learning-based deep learning framework that learns a model of personal graphic design style. PseudoClient operates in an example-based manner, without relying on subjective style terms, and requires only a small handful of examples. In various experiments, we demonstrate that PseudoClient outperforms several alternative methods and offer an understanding of the levers that designers can alter to adjust or improve PseudoClient's ability in learning personal style. Finally, we discuss several applications that could be powered by PseudoClient from three directions: search, feedback, and generation. 

This work takes a step towards the philosophy of computational design understanding with only a small number of data samples, which we argue increases the practicality of machine learning methods for design applications. We hope PseudoClient can be used as a building block to support the development of future design applications and serve as a foundation for future work on computationally understanding visual design style. \rev{For more information, please visit \href{https://chuanenlin.com/personalstyle}{https://chuanenlin.com/personalstyle}.}


\begin{acks}
\rev{We thank all participants who selected images for our experiments and the anonymous reviewers for their comments We also thank Zach Rupert, Tatooine Girl, Ivan Dubovik, Justin Bryant, Aleksandr Reva, Maycon Prasniewski, Hanna Ak, Tyler Thorny, Tatak Waskitho, Kane Young, Rick Barker, Paulius Kolodzeiskis, Anna Shulha, Vee Are, Evan Brown, and Jeff Trish for their beautiful illustrations. This work is supported by a gift from Accenture Technology Labs.}
\end{acks}

\bibliographystyle{ACM-Reference-Format}
\bibliography{base}

% \appendix

% \section{Snapshot of the Participants' Examples}

% \begin{figure}[h]
%   \centering
%   \includegraphics[width=15cm]{figures/snapshot.jpg}
%   \caption{A snapshot of each participants' positive and negative sets. Each column represents a participant. The top row displays an example of their positive set and the bottom row displays an example of their negative set.}
%   \Description{Some sample data for the experiments}
%   \label{fig:snapshot}
% \end{figure}

\end{document}